\documentclass[final]{cvpr}

\usepackage{times}
\usepackage{epsfig}
\usepackage{graphicx}
\usepackage{amsmath}
\usepackage{amssymb}
\usepackage{subfigure}

\usepackage{caption}

\usepackage[pagebackref=true,breaklinks=true,colorlinks,bookmarks=false]{hyperref}

\def\cvprPaperID{} 
\def\confYear{}

\usepackage[T1]{fontenc}
\usepackage[colorlinks]{hyperref}
\makeatletter
\newcommand*{\emailskaist}[2][@kaist.ac.kr]{%
    \def\@tempa{\@gobble}%
    \@for\qrr@email:=#2\do{%
        \edef\@tempb{\noexpand\href{mailto:\qrr@email #1}{\qrr@email}}%
        \edef\@tempa{\unexpanded\expandafter{\@tempa}{, }\unexpanded\expandafter{\@tempb}}}%
    \{\@tempa\}#1%
}

\newcommand*{\emailsitsc}[2][@itsc.kr]{%
    \def\@tempa{\@gobble}%
    \@for\qrr@email:=#2\do{%
        \edef\@tempb{\noexpand\href{mailto:\qrr@email #1}{\qrr@email#1}}%
        \edef\@tempa{\unexpanded\expandafter{\@tempa}{, }\unexpanded\expandafter{\@tempb }}}%
    \@tempa%
}
\makeatother

\begin{document}

\title{Data Augmentation using Random Image Cropping for High-resolution Virtual Try-On (VITON-CROP)}

\author{
	Taewon Kang$^{1}$\thanks{Contributed to the work during his KAIST AI internship},
	Sunghyun Park$^{2}$,
	Seunghwan Choi$^{2}$, and Jaegul Choo$^{2}$\thanks{Corresponding author}\\
	Korea University, Seoul, South Korea$^{1}$ \\ KAIST, Daejeon, South Korea$^{2}$\\
    \emailsitsc{itschool}, \emailskaist{psh01087,shadow2496,jchoo} \\
}

\maketitle

\begin{abstract}
   Image-based virtual try-on provides the capacity to transfer a clothing item onto a photo of a given person, which is usually accomplished by warping the item to a given human pose and adjusting the warped item to the person. However, the results of real-world synthetic images (\textit{e.g.}, selfies) from the previous method is not realistic because of the limitations which result in the neck being misrepresented and significant changes to the style of the garment. To address these challenges, we propose a novel method to solve this unique issue, called VITON-CROP. VITON-CROP synthesizes images more robustly when integrated with random crop augmentation compared to the existing state-of-the-art virtual try-on models. In the experiments, we demonstrate that VITON-CROP is superior to VITON-HD both qualitatively and quantitatively.
\end{abstract}

\section{Introduction}
Currently, online shopping has established a huge commercial advantage compared to in-person shopping, but there are a lot of limitations. Some shortcomings of online shopping include people being unfamiliar with the process, the desire to view products in person, and the unreliability of sizing.

To address these problems, a virtual try-on network (VITON)~\cite{han2018viton,wang2018toward,yu2019vtnfp,lewis2021vogue,yang2020towards,neuberger2020image, dong2019fw} provides product information that is usually obtained through in-person, direct product inspection. This allows the users to experience wearing different clothes without physically changing. Prompted by the speedy development of image synthesis, image-based VITON is aimed at transferring articles of clothing onto a person.

VITON~\cite{han2018viton} uses a coarse-to-fine strategy to seamlessly transfer the desired garment onto the person's corresponding area, without using any 3D information. Image-based VITON follow two processes. In the older process, they warp the image of the clothes to fit the human body, and fuse the image of warped clothing with the person image including pixel-level refinement. Some recent methods added modules that generate segmentation maps to determine the layout of people from the final image.

Recently, VITON-HD~\cite{choi2021viton} proposed to address the low resolution of the synthetic images.  VITON-HD synthesizes $1024 \times 768$ virtual try-on images and is divided into segmentation generation, clothes deformation, and try-on synthesis modules, which allows it to reduce the dependency of the clothing and eliminate misalignment between the warped clothes and the corresponding clothing regions.

However, the result of the real-world synthetic images (\textit{e.g.,} selfies) from the previous method does not look realistic. For instance, in a real situation using virtual try-on, input images that do not include part of the upper and lower body may be used for VITON. We define selfies as an example of a real-world VITON problem.

There are two shortcomings with the previous VITON: the segmentation problem and the cloth problem. The segmentation problem is caused by an error in the segmentation generator which results in the gap between the face and the body being distorted, and the neck being misrepresented. The cloth problem is in reference to the issue of wrinkles on the clothing, excessive fabric on certain areas, the deletion of parts of the garment, and/or significant changes to the style of the garment. To address these problems, we propose a novel method called VITON-CROP.

VITON-CROP is an image-based virtual try-on model with random crop augmentation which allows cropped images to be used for virtual try-on tasks regardless of the aforementioned barriers. In the experiments, we demonstrate that VITON-CROP is the superior method compared to existing state-of-the-art virtual try-on models.

\begin{figure*}[t]
    \captionsetup[subfigure]{}
	\begin{center}
	\renewcommand{\thesubfigure}{}
	\subfigure[]
	{\includegraphics[width=0.165\linewidth]{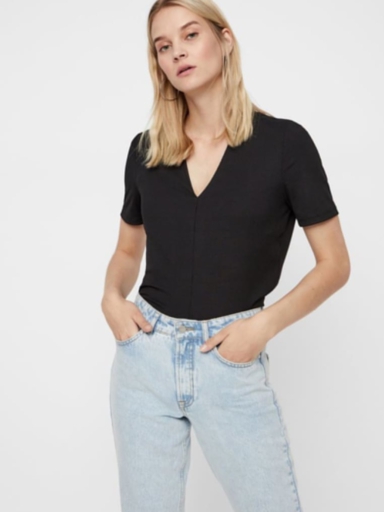}}\hfill
	\subfigure[]
 	{\includegraphics[width=0.165\linewidth]{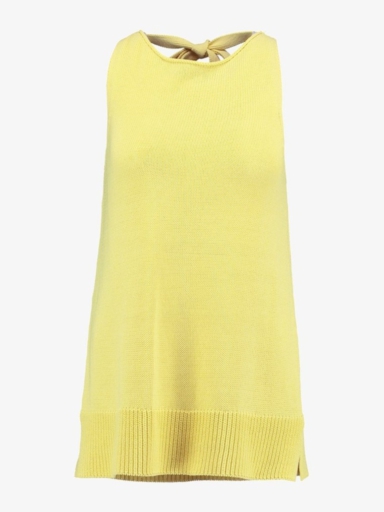}}\hfill
 	\subfigure[]
 	{\includegraphics[width=0.165\linewidth]{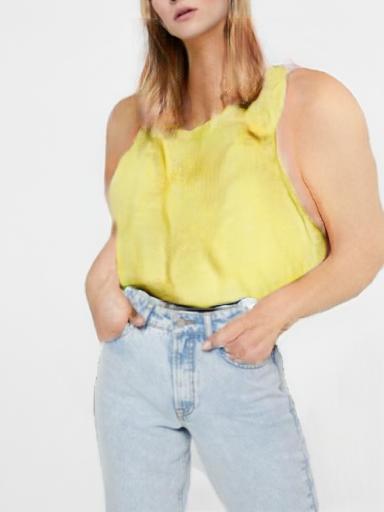}}\hfill
 	\subfigure[]
 	{\includegraphics[width=0.165\linewidth]{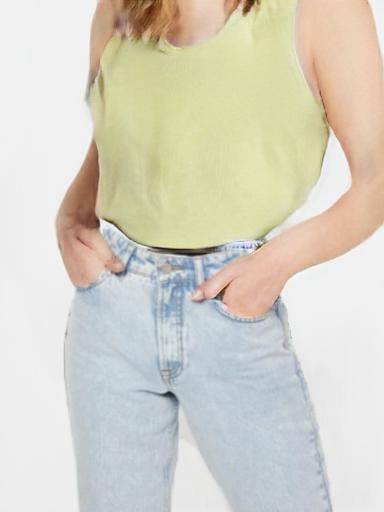}}\hfill
 	\subfigure[]
 	{\includegraphics[width=0.165\linewidth]{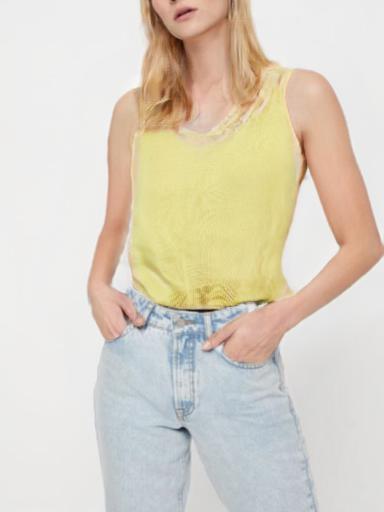}}\hfill
 	\subfigure[]
 	{\includegraphics[width=0.165\linewidth]{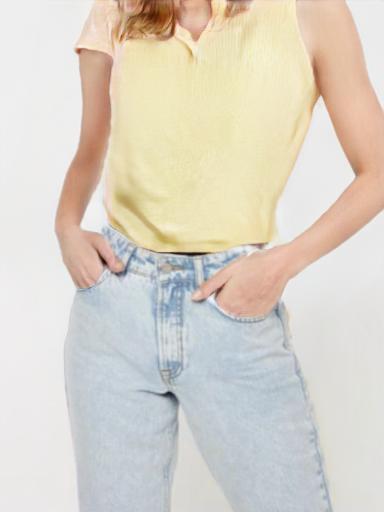}}\hfill\\
 	\vspace{-22pt}
 	\subfigure[Reference Image]
	{\includegraphics[width=0.165\linewidth]{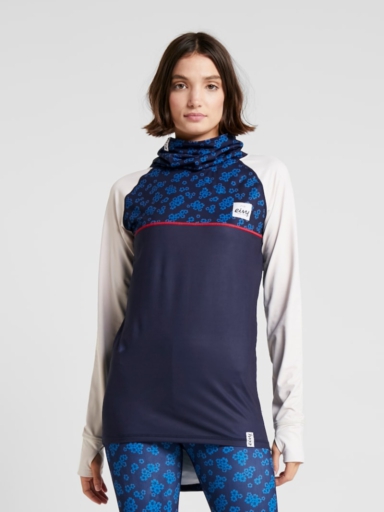}}\hfill
	\subfigure[Target Clothes]
 	{\includegraphics[width=0.165\linewidth]{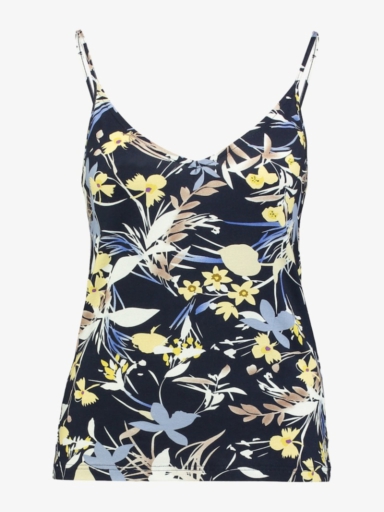}}\hfill
 	\subfigure[VITON-HD (scale=0.7)]
 	{\includegraphics[width=0.165\linewidth]{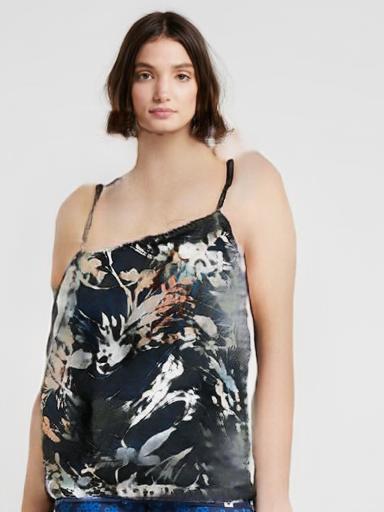}}\hfill
 	\subfigure[VITON-HD (scale=0.5)]
 	{\includegraphics[width=0.165\linewidth]{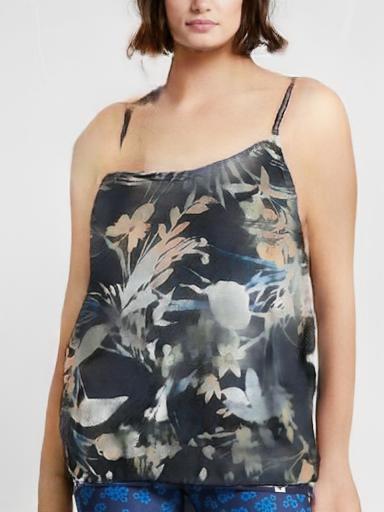}}\hfill
 	\subfigure[Ours (scale=0.7)]
 	{\includegraphics[width=0.165\linewidth]{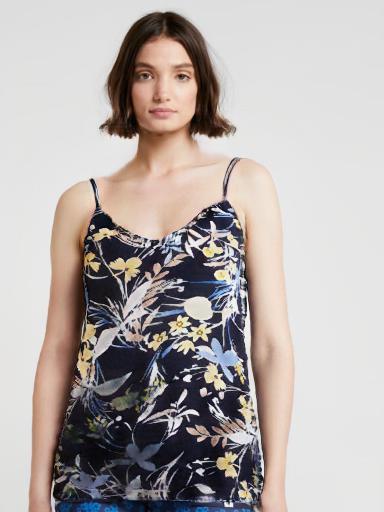}}\hfill
 	\subfigure[Ours (scale=0.5)]
 	{\includegraphics[width=0.165\linewidth]{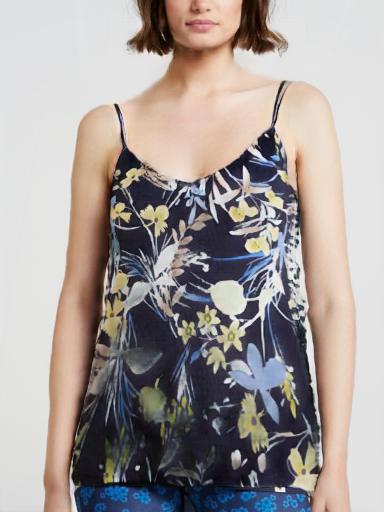}}\hfill\\
    \end{center}
    \vspace{-15pt}
    \caption{\textbf{Qualitative result:} Qualitative comparison of the VITON-HD\cite{choi2021viton} method. Given a pair of a reference(containing a person) and a target clothing, our method successfully generates cropped images to be used for virtual try-on tasks.}\label{Fig01}
\end{figure*}

\section{VITON-CROP}

Given a reference image $I$ of a person and a clothing image $c$, the goal of the original VITON is to generate a synthetic image $I$ of the same person wearing the target clothes $c$, where the pose and body shape of $I$ and the details of $c$ are preserved.

Similar to the implementation of VITON-HD \cite{choi2021viton}, we use $(I, c, I)$ tuples where the reference image $I$ figure presents in clothing image $c$. Since direct usage of the procedure on $(I, c, I)$ can harm the model’s generalization ability during testing, we first generate a clothing-agnostic person representation that leaves out the $c$ information and uses it as an input. There are three different models: (1) Segmentation Generator, (2) Clothes Deformation, (3) Try-On Synthesis.


During the three processes of VITON-HD, we found the aforementioned limitations of VITON-HD which contains the neck being misrepresented and significant changes to the style of the garment. So, we propose a VITON-CROP model that allows cropped images for virtual try-on tasks including extreme conditions involving asymetrical areas and fine details that are typically hard to align, which we will refer to as micro-scale-cropped datasets.

Combined with simple random crop augmentation, the above three models can synthesize images more robustly, accurately and precisely compared to VITON-HD. In this paper, our model architecture is devised by combining several random crop augmentation modules into three modules of the VITON-HD.

\section{Experimental Results}
\subsection{Experiment Setup}
We follow the default setting of VITON-HD with edited discriminator loss and we use a 1024×768 virtual try-on dataset (zalando dataset, crawled by VITON-HD\cite{choi2021viton}) for the purpose of VITON-HD research since the resolution of images on the dataset provided by existing models is low.


We use pairs of people and clothing images to evaluate paired settings and we mix the clothing image to evaluate an unpaired setting. In order to reconstruct the person’s image with the original clothing item we use the paired setting, and the unpaired setting to change the clothing item of the person’s image to another item.

During the training process (pair setting), we crop the dataset in the code using the PyTorch RandomResizedCrop \footnote{We follow the procedure of RandomResizedCrop as implemented in the PyTorch transform \cite{paszke2019pytorch} \url{https://pytorch.org/vision/stable/_modules/torchvision/transforms/transforms.html} } library, which is very similar to the augmentations used by other toolkits like Caffe and the original AlexNet.

RandomResizedCrop takes as input an $H \times W$ image, selects an area at random with consider of a random aspect ratio, and resizes the image to output a $H_{resized} \times W_{resized}$ crop. During the test process (unpaired setting) we pre-crop the dataset using the RandomResizedCrop script.

\begin{figure}[t]
    \captionsetup[subfigure]{}
	\begin{center}
	\renewcommand{\thesubfigure}{}
 	\subfigure[scale=1.0]
	{\includegraphics[width=0.33\linewidth]{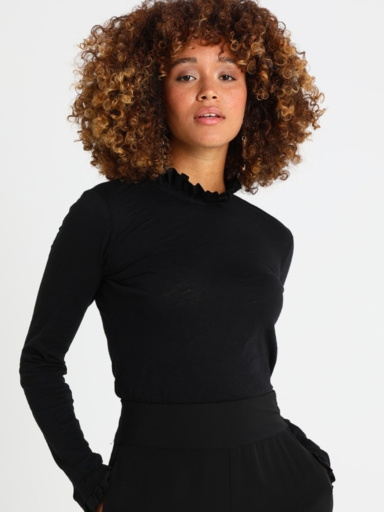}}\hfill
	\subfigure[scale=0.7]
 	{\includegraphics[width=0.33\linewidth]{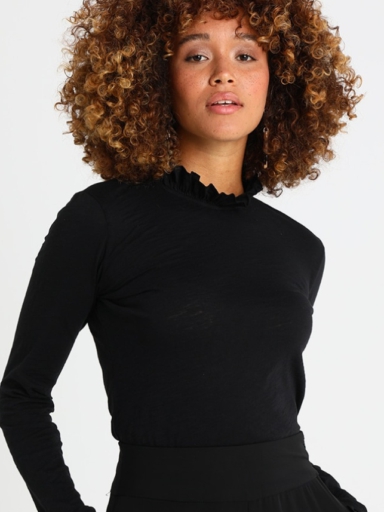}}\hfill
 	\subfigure[scale=0.5]
 	{\includegraphics[width=0.33\linewidth]{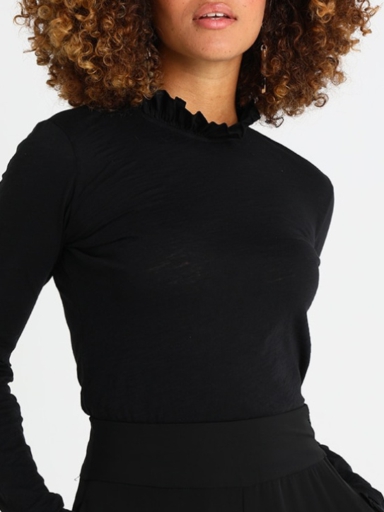}}\hfill \\
    \end{center}
	\caption{\textbf{Example of scale parameters:} We crop the dataset using scale parameters, this picture shows the example usages of scale parameters.}
	\label{fig_crop_ex}
\end{figure}

In addition, we crop the dataset using scale parameters, such as scale=0.7, scale=0.5, etc.  All scale ranges are 0.5 to 1.0 in the training process, and scale=0.7 denotes the scale range is 0.7 to 0.7 during the testing process. Detailed example of scale parameters are seen in Fig.~\ref{fig_crop_ex}. Both training and test process datasets are resized to $512 \times 384$.

\subsection{Qualitative Analysis}

We compare VITON-CROP with VITON-HD, using the official implementation. We follow the training and testing procedure of VITON-HD. For the comparison with baselines, our model has a quantitatively large effect, and the detailed results are seen in Fig.~\ref{Fig01}.

\subsection{Quantitative Analysis}

We perform the quantitative analysis in an unpaired setting, in which a person wears the new clothes. We evaluate our method using metrics commonly used in virtual try-ons. We use the fréchet inception distance (FID)\cite{heusel2017gans} score in the unpaired setting.

\begin{figure}[ht]
\begin{center}
\includegraphics[width=7cm]{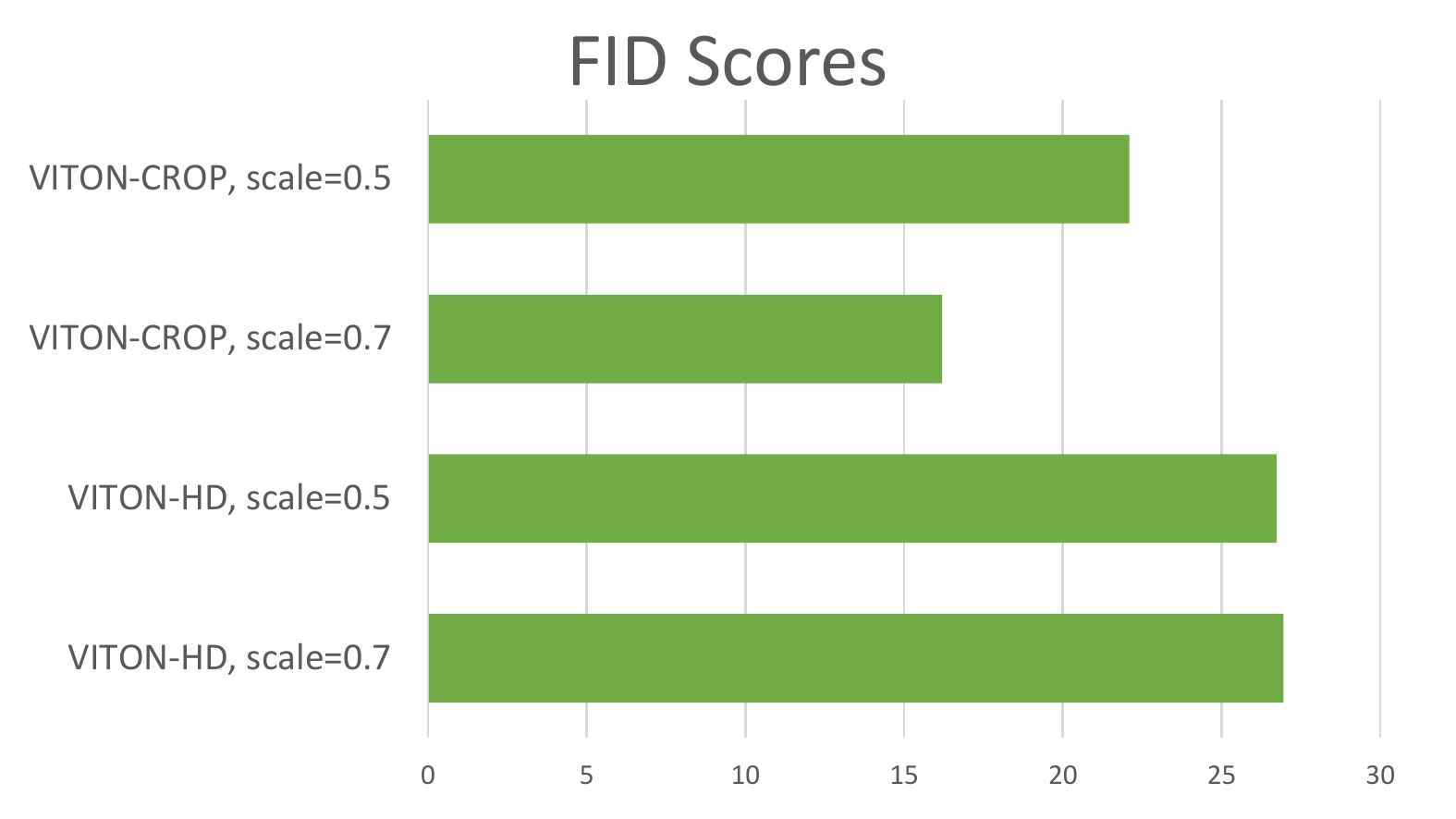}
\end{center}

\caption{\textbf{Quantitative result:} We use the fréchet inception distance (FID) score to measure the performance of VITON-HD, with the same zalando-hd dataset. This figure shows that our crop augmentation has a quantitatively large effect compared to VITON-HD.}\label{Fig02}
\end{figure}

The inception score is not included in the experiments, since it cannot measure whether the details of the clothing image are maintained. We confirm that crop augmentation has a quantitatively large effect in VITON-HD and the detailed analysis can be seen in Fig.~\ref{Fig02}.

\section{Conclusions}

This paper demonstrated that our proposed approach called VITON-CROP well synthesizes examples of real-world synthetic images and crop datasets. We confirm that crop augmentation in VITON shows outstanding performance which allows cropped images for virtual try-on tasks with extreme conditions referred to as micro-scale-cropped datasets. In the experiments, we demonstrate that VITON-CROP has superiority compared to existing state-of-the-art virtual try-on tasks both qualitatively and quantitatively.

{\small
\bibliographystyle{ieee_fullname}
\bibliography{egbib}
}

\end{document}